# Unsupervised Low-Dimensional Vector Representations for Words, Phrases and Text that are Transparent, Scalable, and produce Similarity Metrics that are Complementary to Neural Embeddings

Neil R. Smalheiser* and Gary Bonifield

Department of Psychiatry and Psychiatric Institute, University of Illinois College of Medicine,

Chicago, IL 60612 USA *neils@uic.edu

## ABSTRACT

Neural embeddings are a popular set of methods for representing words, phrases or text as a low dimensional vector (typically 50-500 dimensions). However, it is difficult to interpret these dimensions in a meaningful manner, and creating neural embeddings requires extensive training and tuning of multiple parameters and hyperparameters. We present here a simple unsupervised method for representing words, phrases or text as a low dimensional vector, in which the meaning and relative importance of dimensions is transparent to inspection. We have created a near-comprehensive vector representation of words, and selected bigrams, trigrams and abbreviations, using the set of titles and abstracts in PubMed as a corpus. This vector is used to create several novel implicit word-word and text-text similarity metrics. The implicit word-word similarity metrics correlate well with human judgement of word pair similarity and relatedness, and outperform or equal all other reported methods on a variety of biomedical benchmarks, including several implementations of neural embeddings trained on PubMed corpora. Our implicit word-word metrics capture different aspects of word-word relatedness than word2vec-based metrics and are only partially correlated (rho = ~0.5-0.8 depending on task and corpus).

The vector representations of words, bigrams, trigrams, abbreviations, and PubMed title+abstracts are all publicly available from http://arrowsmith.psych.uic.edu for release under CC-BY-NC license. Several public web query interfaces are also available at the same site, including one which allows the user to specify a given word and view its most closely related terms according to direct co-occurrence as well as different implicit similarity metrics.

# INTRODUCTION

A recent lawsuit in the United Kingdom involved a patient suing a hospital over access to his health records. The patient felt that he had been subjected to unnecessary surgery, and sought access to all records related to his care. It developed that the hospital had supplied him with only a subset of the relevant information, relying on a technical distinction between the words *notes* and *records*. It is a good example of what the forensic linguist John Olsson has identified as a common tactic of scuzzy businesses: relying on technical distinctions between similar words to do something or other. But, what does it mean for two words to be "similar"? It seems obvious, but this turns out to be a vexing question--and one whose answer has implications for many fields of inquiry, ranging from psycholinguistics (where it is studied in the context of exploring hypotheses about how our brain recognizes words) to biomedical ontology (where it has implications for doing inference over the output of high-throughput assays) to natural language processing and text mining.

One reason that word similarity is difficult to study is that two words can be similar to each other in many different ways. We can divide these ways of being "similar" into a number of broad categories, ranging from relationships of abstract meaning (e.g. *egg* and *ovum*) based in large part on human intuitions, to purely statistical relationships derived from large sets of linguistic data (e.g. the word *gene* tends to be joined by the word *and* to the words *locus* and *region* more frequently than would be expected by chance, while the word *protein* is more likely to be joined with the word *enzyme*). Two words may be related, such as "pork" and "beans". Or they may co-occur in a phrase, such as "Cold Spring". Two words may also be related insofar as they can substitute for each other within a given utterance, as the word "cat" in "The cat is on the mat" can be substituted by a wide number of other entities that can sit or stand on a mat. Furthermore, two words may be implicitly related if they share some relation with a third entity; for example, "acupuncture" and "morphine" are both treatments for pain. In the present study, we describe and characterize several similarity metrics for relating words, terms, or passages of text to each other. We define direct similarity of two terms as measuring how often they co-occur within the same text, relative to the frequency expected by chance. Implicit term or text similarity metrics measure the degree to which two terms or texts share similar vector representations (see below). Both direct and implicit types of similarity have their place, depending on the intended purpose, and indeed, having a palette of text-related similarity metrics should be a boon for text mining models.

Many previous investigations have examined different approaches to estimating the similarity of two words or phrases, concepts, as well as sentences and documents. To give just a few examples, two words can be related according to their path distance on graphs based on network features, ontologies, or Wordnet [1]. Alternatively, they can related by direct measures of text similarity such as string matching [2], or by implicit measures such as the number of shared words in their dictionary definitions [3]. Two articles or documents can be judged for similarity of their text words, their metadata (e.g., number of shared MeSH terms, MeSH term pairs [4] or UMLS concepts [5]), or implicit information such as similarity in their cited or citing articles [6]. There are various unsupervised schemes for representing documents in terms of underlying word similarities (among them latent semantic indexing [7], PubMed Related articles [8] and neural vector spaces [9]), which can be used to create word-word or topic-based document-document similarity metrics that rank articles for relatedness (see also [10, 11]).

During the past five years, neural embedding methods such as word2vec [12] and GloVE [13] have been investigated widely to create low-dimensional vector representations of words and text passages. In such schemes, the implicit similarity of any two words or texts is given by the cosine similarity of their vectors [e.g., 9, 14]. Despite the popularity of this approach, there are several issues that should be acknowledged. First, the most popular implementations use a sliding window that relates a central word to its nearby contextual words, and this tends to emphasize word substitutionality. Second, the vectors vary according to the type and size of the textual corpus used to train the models; typically, a very large corpus of text is used that may not be the same as the text to which the modeling is being applied. Third, there are many different variations and implementations of neural embeddings; not only are there many parameters and hyperparameters that need to be tuned to optimize any particular modeling task, but the optimal tuning will vary from one specific task to another. Fourth, typically a word is represented by a vector having 50-500 dimensions, whose values are assigned by a neural network operating as a black box. The interpretation and relative importance of each dimension are not transparent to interpretation, at least not without extensive further training and processing [15, 16].

To provide an alternative type of similarity metric, we created a simple unsupervised method for representing words, phrases or text as a low dimensional vector, in which the meaning and relative importance of dimensions is transparent to inspection. The approach is general, although we have employed PubMed titles and abstracts as a corpus in order to apply the metrics to biomedical tasks. Our implicit term metrics give superior performance on a panel of biomedical term similarity and relatedness benchmarks. We also demonstrate how our metrics compare with several direct similarity metrics and word2vec-derived metrics for computing similarity of PubMed text passages (title+abstracts).

The present studies are part of an ongoing project to create a series of implicit similarity metrics to aid in text mining and modeling of the biomedical literature. For example, recently we created journal similarity metrics that relate any two journals A and B according to a) how similar they are topically, b) how similar they are in terms of sharing the same authors, and c) how likely it is that an author who publishes in A will publish later in B [17]. We also created MeSH (Medical Subject Headings) similarity metrics that relate any two MeSH terms A and B according to a) how often the two terms co-occur in a single article, and b) how often the two terms are found within the body of articles published by a single author [18]. Such implicit metrics are valuable in text mining models such as are used in the Author-ity author name disambiguation project [19, 20], where the goal is to consider two articles that share the same author [lastname, first initial] and estimate the probability that the two articles refer to the same author-individual. A match on (say) journal name is a powerful direct feature suggesting that the two articles may share the same author, but this feature can only be scored as nonmatch (= 0) vs. match (= 1). In contrast, two nonmatching journal names can be scored using implicit similarity metrics as a positive number between 0 and 1 in all cases. Thus, implicit features have a "smoothing" effect, improving robustness and sensitivity of sparse models.

**MATERIALS AND METHODS**

**Terms.**

A database of all PubMed articles published 1966-2016, in English (or with titles and abstracts translated into English), was created. Titles and abstracts were tokenized as in [21] and stoplisted using the PubMed 365-word stoplist [22]. Words were included only if they appeared in at least 100 titles and in at least 25 abstracts. The **basic implicit word model** consists of 44,201 words that met these criteria and includes 41,918,680 word pairs that co-occurred in the title or abstract of the same PubMed article.

The **Direct Odds Ratio** provides a normalized measure of co-occurrence frequency of two words or terms. That is, it measures how often two words are observed to co-occur in the same article, relative to the value that would be expected by chance (i.e., if each word occurred at random throughout the corpus of PubMed articles and independently of each other). The two terms may co-occur in either order, anywhere in the same article, and need not be within the same sentence or to within a proximity window of defined size. To compute the Direct Odds Ratio, the geometric mean (geomean) of document frequency of each pair of co-occurring terms was computed, and the output was stored in a file sorted by ascending geomean.

$$\text{geomean termA:termB} = \sqrt{\text{termA doc frq} * \text{termB doc frq}}$$

This geomean output file was then processed in bins of 500 to compute the odds ratio for each pair of co-occurring terms where

$$\text{Direct odds ratio} = \text{doc co-occurrence frq} / \text{bin average}.$$

The bin average has been used to estimate the expected chance level of co-occurrence in several of our previous metrics investigations [1, 2]. Note that the direct odds ratio is not only employed to create vector representations of words and terms (as discussed below), but also provides a **direct similarity metric**. That is, two words or terms show greater direct similarity if they tend to co-occur in the same articles more than would be expected by chance.

For the **full model**, we included selected bigrams, trigrams, and abbreviations as follows: 1. Bigrams were formed from tokenized words that appear in the basic model. 2. The two words making up the bigram must have a direct odds ratio of 10 or greater. 3. Trigrams are formed from bigrams where trigram ABC is derived from bigrams AB and BC. Trigrams must appear in 1000 or more abstracts, and the trigram ABC must have a document frequency at least 30% of the minimum document frequencies of AB and BC.

Note that when computing multi-word phrases, instances of the individual words that make up these bigrams or trigrams were removed from further consideration, and were <u>not</u> included when computing direct and implicit word metrics of the individual words.

To select abbreviations for encoding by their own vectors: 1. We only considered instances of abbreviations that are written in all-capitals and listed in at least 25 articles within the ADAM abbreviation database [23, 24]. 2. The frequency of the non-capitalized word must be greater than the frequency of the all-capitalized word. Again, when an abbreviation was selected, instances of the abbreviation were excluded when computing the same token considered as a single non-capitalized word. For example, ACT is an abbreviation whereas act is a non-capitalized word; instances of ACT were not included when computing the vector for act.)

The **full implicit term model** is made up of 97,754 terms (43,494 words, 49,918 bigrams, 3,638 trigrams, and 704 abbreviations) and includes 92,757,119 co-occurring term pairs.

**Computing vector representations and implicit similarity metrics for terms.**

For each term selected as above, we made a list of all other selected terms that co-occurred with it in titles and abstracts of MEDLINE articles, and ordered the list according to their direct odds ratios. The co-occurring terms with the highest direct odds ratios were used to form a vector. That is, for a given term, we took its co-occurring term having the highest direct odds ratio and placed it in dimension 1; took the term having the next highest direct odds ratio and placed it in dimension 2; and so on, until 300 dimensions were assigned or until the direct odds ratio fell below 1.25. **Thus, the vector representation of a word or term consists of a ranked list of its co-occurring context terms.**

We chose the cut-off criteria (300 dimensions or direct odds ratio of 1.25) because of previous research suggesting that 300 dimensions gives asymptotically optimal performance in both neural embedding models [e.g., 19] and latent semantic indexing [11]. We also felt that including terms having a direct odds ratio of less than 1.25 would be too close to random co-occurrence, hence might subject the vector to too much "noise". The vast majority of single words (99.9%) in the basic model, and 97% of terms in the full model, had at least 300 co-occurring terms whose direct odds ratio was 1.25 or greater. (The few exceptions were mainly multi-word terms of very low frequency.)

Creating implicit similarity metrics for terms:

a) **Implicit Shared Terms**. For any two terms in our dataset, we examined their 300-dimensional vectors, and counted the number of words (or terms) shared in both vectors. The unweighted similarity is thus an integer ranging from 0 to 300.

b) **Implicit Weighted Score for the basic (single word) model**. For two words or terms A and B, we examined their 300-dimensional vectors, and list the words shared in both vectors. We then created a weighted sum as follows: For each shared vector word, choose the LESSER of the odds ratio in the two vectors. Then, the implicit weighted score = the sum of the $\log_e$ odds ratios across all shared words.

c) **Implicit Weighted Score for the full (term) model**. We give greater weight for shared trigrams and bigrams relative to shared words or abbreviations. Thus, we create a weighted sum as follows: For each shared vector term, choose the LESSER of the direct odds ratio in the two vectors. Then, the implicit weighted score = the sum of ($w_i$ * $\log_e$ odds ratios) across all shared words, where $w_i$ = 1 for words and abbreviations, 2 for bigrams, 3 for trigrams.

**Computing vector representations of text passages.**

For each PubMed article, the title+abstract was concatenated to form a single text passage, and tokenized and processed to recognize words, bigrams, trigrams, abbreviations from our dataset. Each term was only counted once, i.e., there was no double-counting for the title or for multiple instances in the abstract. For each term found in the title+abstract, its 300-dimensional vector representation was listed. Then, we created a master list of all terms that occur across all these vectors. Each term was assigned a value which is the sum of the log$_e$ direct odds ratio of that term in each vector it appears in. Finally, we represented the PubMed title+abstract as a 300-dimensional vector as follows: The term having the highest value gets rank 1, next highest is rank 2, etc. down to 300 dimensions.

**Computing implicit similarity metrics for text passages.**

For two articles (title+abstract) or other text passages, we examine their 300-dimensional vector representations, and listed the words (or terms) shared in both vectors. The **implicit shared terms** metric is simply the number of shared terms, i.e., an integer between 0 and 300. The **implicit weighted score** is computed as follows: For each shared vector term, choose the LESSER of the direct odds ratio in the two vectors. Then, the implicit weighted score = the sum of ($w_i$* log$_e$ odds ratios) across all shared terms, where $w_i$ = 1 for words and abbreviations, 2 for bigrams, 3 for trigrams.

**Word2vec based similarity metrics.**

For computing word2vec term similarity, we used the University of Turku word2vec 200-dimensional vectors computed for biomedical words that was trained on PubMed titles and abstracts downloaded from http://evexdb.org/pmresources/vec-space-models/. To represent phrases, some bigrams were already represented as single entities in the Finnish word2vec corpus. However, we verified that the performance of word2vec on the 30 term relatedness benchmark (Supplemental File 1) was improved by summing individual word vectors for multi-word phrases [16, 25], thus summing was performed on the other benchmarks as well. This vector representation is referred to here as "**word2vec**". To obtain the word2vec similarity of two words or phrases, the cosine similarity of their word2vec vectors was computed, giving a real number between -1 and +1.

To represent PubMed articles, word2vec 300-dimensional vectors were computed using the paragraph2vec code https://github.com/hassyGo/paragraph-vector using the following parameters: -wvdim 300 -pvdim 300 -itr 10 [25, 26]. Training was performed across the entire PubMed corpus of article titles and abstracts (1966-2016). This training procedure created a separate word2vec representation of biomedical words which will be referred to as "**pvtopic**" since it follows the word2vec vector representation described in Hashimoto et al. [26].

Although Word2vec and pvtopic are both implementations of neural embeddings based on a PubMed corpus, they differ in performance on the 30 term relatedness benchmark (see Supplemental File 1), so word2vec was used for term similarity and relatedness comparisons, whereas pvtopic was used for representing PubMed articles in order to follow the method of Hashimoto et al [26]. The paragraph2vec code computed 300-dimensional vector representations for all PubMed articles; each article was represented by concatenating its title with its abstract to form one text. To obtain the similarity of two text passages, the cosine similarity of their vectors were computed. To ensure robustness, similarity scores were only computed for articles that contained abstracts and whose title+abstract contained at least 25 words.

**Processing of biomedical term relatedness and term similarity benchmarks.**

Several biomedical term relatedness and term similarity benchmarks were downloaded from http://rxinformatics.umn.edu/SemanticRelatednessResources.htm and shown in Supplemental files 1-5. Most of the words and phrases listed in the benchmarks had exact matches to our dataset of words and phrases used for calculating direct and implicit similarity metrics. When exact matches were not found, terms were manually mapped to the closest words or multi-word terms in our dataset. (For example, some brand names for drugs were mapped to their generic counterparts.) For the words that failed to map to any terms in our dataset, the involved term pairs were removed from consideration. Similarly, for the words that failed to map to the word2vec corpus, the involved term pairs were removed from consideration when computing word2vec similarity. Term mappings to our dataset are shown in Supplemental Files 1-5.

In order to assess whether the Spearman rank correlations showed statistically significant differences for different metrics, we divided the benchmark dataset randomly into five equal subsets and computed the Spearman correlations for each subset. This gave a mean, SD, and SEM for each metric, and allowed us to compare different metrics by unpaired, two-tailed t-test using an online t-test calculator (https://www.graphpad.com/quickcalcs/ttest1.cfm).

**PubMed article record pairs that match (or do not match) on author.**

Using the 2009 Author-ity name disambiguation dataset [3, 4, 35], we randomly chose 1,000 pairs of sole-author articles published in 1987-2009 in English that had abstracts (and the total number of words in title+abstract was $\geq$25) that were predicted to be written by the same individual, vs. 1,000 pairs of sole-author articles that were predicted to be written by different individuals (i.e., sharing the same last name but having a mismatch on the first initial of the author's name). Pairs were excluded if they had identical titles or abstracts. For each pair of articles, we computed the following six similarity measures on the title+abstract. Direct similarity measures included a) number of shared title words (after stoplisting, tokenization and stemming), b) longest common character string (of at least 3 characters, including spaces), and c) number of shared rare terms (i.e., terms that occurred in fewer than 25 abstracts in MEDLINE, weighted by counting shared bigrams as 2 and shared trigrams as 3). We also examined three implicit similarity measures: d) implicit shared terms, e) implicit weighted score, and f) pvtopic word2vec-based similarity score.

**RESULTS**

**a) Comparison of metrics for term similarity of selected biomedical words**

The direct similarity of two words or terms measures how often the two co-occur in the same article, relative to the co-occurrence that would be expected simply from the document frequencies of the two considered independently. For example, "peanut" and "butter" co-occur often (because of the phrase "peanut butter") and thus show high direct similarity, i.e., high direct odds ratio. In contrast, the implicit similarity of two words or terms measures how well they share the same context words. For example, "randomization" and "randomisation" have very low direct similarity since they rarely co-occur in the same article (one is American spelling and one is British spelling), but they share many of the same surrounding context words (e.g., "clinical trial" and "RCT") and so have very high implicit similarity. Finally, the word2vec metrics measure the substitutability of one word or term for another within passages of text. For example, in the sentence "I took the train from Rome to Florence", the word "I" could be readily substituted by another personal name (e.g., "he", "she", "Sheldon", etc.), the word "train" could be substituted by other modes of transportation (e.g., "bus" or "bike"), and "Florence" could be substituted by the name of another city that can be reached by train (but less likely by a distant city such as Chicago).

To give an example of how the different metrics rank related words most highly in the PubMed corpus, Table 1 shows the top 10 most related words to "tennis" by direct odds ratio, by implicit weighted score, and by word2vec. The direct odds ratio highlights words such as "player", "elite", "elbow", and "epicondylitis" which are described in articles that study tennis players. In contrast, 9 of the top 10 words by the word2vec metric are names of other sports, such as basketball and baseball. The implicit weighted score falls in between the other two metrics, sharing 3 words with the direct odds ratio and 5 words with word2vec. Only a single word, "soccer", is shared in the top ten by all three metrics.

To give another illustrative example, Table 2 shows the top 10 most related words to "Italy". The top 10 words by direct odds ratio consists entirely of the names of cities and regions within Italy. In contrast, top 10 words according to the word2vec metric consist entirely of the names of other European countries. Again, the implicit weighted score falls in between the other two, as it highlights words that are used often in the same context as the word Italy, including words such as "Italian", "Europe", "European" and "Mediterranean", as well as the names of five other European countries. These examples confirm our expectation that the direct, implicit and word2vec metrics are capturing different types of similarity, at least in part.

**b) Performance of metrics on a 30 term biomedical term relatedness benchmark.**

We compared how well various word similarity metrics align with the mean judgments of three physician and nine medical coders of term relatedness (rated on a scale of 1 to 4) in the Pedersen 30 word pair benchmark [5]. These term pairs are a subset of the larger Pakhomov 101 term pair dataset (see below), chosen because they showed high inter-rater reliability.

The 30 word benchmark, physician and coder relatedness scores, and similarity scores computed by the various metrics are shown in Supplemental File 1. As shown in Table 3, the implicit shared terms (both weighted and unweighted, computed from the full term model) gave high alignment with human judgments of term similarity, as assessed by Spearman rank correlation. The correlations were extremely high (0.86 for physicians, 0.81 for coders), which to our knowledge exceeds that previously reported for any term similarity metrics tested where the pair of terms is the sole input to the metric (i.e., excluding approaches which employ supplementary outside information from knowledgebases) [e.g., 5, 9, 27, 28]. The direct odds ratio also gave high Spearman correlations (0.82-0.84). Of several different ways examined to compute the word2vec and pvtopic similarities (see Supplemental File 1), the best performance obtained for word2vec was 0.79 for physicians, 0.80 for coders, and the best Spearman rank correlation of pvtopic was 0.676 for physicians, 0.69 for coders. These values are comparable to or better than the performance of other word2vec-based metrics reported by others on this benchmark.

**c) Benchmark of 101 biomedical term pairs rated for relatedness.**

Because the 30 word benchmark is relatively small, we further evaluated our metrics on the larger set of 101 term pairs that were manually rated for relatedness on a scale of 1 to 10 by 13 medical coders, consisting mostly of disease names and signs and symptoms of disease (Pakhomov et al. (2011) [29]). See Supplemental File 2 for the dataset (note that one term pair failed to map to our dataset, giving 100 term pairs).

The performance of our metrics is shown in Table 4. The Spearman rank correlations with coder relatedness judgment were similar for the direct odds ratio (0.7582) and the two implicit metrics (implicit shared terms (0.7426) and implicit weighted score (0.7354)). In contrast, word2vec showed a much lower correlation with human judgment (0.4958) and this difference was statistically significant compared to the implicit weighted score ($p = 0.027$, unpaired, 2-tailed t-test using values derived from 5-fold cross-validation).

Examining the Spearman correlations among the different metrics themselves is a way of assessing whether they are redundant or measure different aspects of similarity and relatedness in the context of a given task or situation. In this dataset, the direct odds ratio was highly correlated with the implicit metrics (>0.85) but showed only a modest correlation with word2vec (0.55) and the implicit metrics also showed modest correlations with word2vec (0.68) (Table 4). These cross-metric correlations are even lower than was observed for the 30 term pair benchmark (Table 3), and confirm again that our implicit metrics are not redundant with word2vec.

**d) Performance on medical resident judgments of biomedical term similarity and relatedness**

The UMNSRS-Similarity benchmark of 566 term pairs involves diseases or conditions and includes many drug names (Pakhomov et al. (2010) [30]). The dataset, which consists mostly of single words, was rated for similarity by medical residents and is shown in Supplemental File 3 (note: we only evaluated the 501 term pairs that mapped to our dataset). The performance of the various metrics is shown in Table 5. Again, the correlations with human judgments of word similarity for direct odds ratio (0.70) and implicit weighted score (0.69) were high and not significantly different from each other ($p = 0.4676$), whereas the performance of word2vec (0.58) was significantly worse than the other metrics (implicit weighted score was significantly greater than word2vec at $p = 0.0169$, unpaired, 2-tailed t-test using values derived from 5-fold cross-validation).

Finally, two versions of the UMNSRS-Relatedness benchmark [30] were examined; the full version containing 588 term pairs (of which 511 term pairs mapped to our dataset), and a modified version consisting of 458 term pairs that occur across different domains (Pakhomov et al. (2016) [31]), of which 420 term pairs mapped to our dataset. As shown in Supplemental Files 4 and 5, the benchmarks consist mostly of single words. In the 511 word benchmark (Table 6), the Spearman rank correlations with human judgments of word relatedness were 0.63 for direct odds ratio and 0.60 for implicit weighted score; these correlations were not significantly different (direct odds ratio vs. implicit weighted score p = 0.3964).

In contrast, word2vec gave significantly lower performance (0.48) than the other metrics (implicit weighted score vs. word2 vec p = 0.0224), and similar to word2vec performance reported by Muneeb et al [32] and Chiu et al [33]. In the 420 word version (Table 7), all metrics gave slightly better performance, but again, word2vec was significantly lower than the other metrics (implicit weighted score vs. word2vec p = 0.0015). The comparison of our implicit metrics against word2vec is fair insofar as neither metric was deliberately tuned for the tasks at hand; however, it should be acknowledged that custom task-specific tuning of the word2vec parameters can improve performance to the point where both metrics are comparable: Henry et al. [28] and Zhu et al. [34] have reported Spearman correlations of 0.64-0.68 across various word2vec models against UMNSRS-Similarity and 0.59-0.72 against UMNSRS-Relatedness.

Across all benchmarks, the direct odds ratio exhibits a relatively high Spearman rank correlation with the implicit metrics (0.79 – 0.91), but a consistently lower correlation with word2vec (0.55 – 0.75). The implicit metrics show a substantial but partial correlation with word2vec (0.68-0.83).

### e) PubMed article record pairs that match (or do not match) on author.

Besides evaluating the similarity metrics on biomedical term similarity and relatedness tasks, we evaluated them for their ability to compare two text passages. Texts can be compared for similarity (or relatedness) in different ways. For example, in information retrieval, one text may be given as a query and other texts may be ranked in terms of their topical relevance. Here, we compared two title+abstract fields of sole-authored PubMed articles and evaluated how well the similarity metrics could distinguish article pairs written by the same individual, vs. written by different individuals. This task has a clear, objective endpoint and is of practical value, since the similarity metrics are potential features to be used in modeling author name disambiguation [3, 4] in cases where articles are sole-authored and therefore presumably written by the individual listed as author. We took 1,000 randomly chosen article pairs that were predicted to be written by the same individual (the positive sample), and 1,000 article pairs that were written by a single author but written by a different individual (i.e., sharing the same last name but a different first initial, which serves as the negative sample). We computed three measures of direct similarity -- number of shared title words, number of shared rare terms, and longest common character string -- as well as three implicit similarity measures, the implicit shared terms, implicit weighted score, and pvtopic (see Methods). The article pairs and computed scores are shown in Supplemental File 6.

The metrics were first assessed in terms of their coverage, that is, the number of article pairs that received nonzero similarity scores, and in terms of their discrimination ratio, that is, the mean similarity value seen across the positive sample divided by the mean value of the negative sample. As shown in Table 8, all of the examined metrics were significantly different (positive vs. negative sample) at p<0.0001 (2-tailed t-test, unpaired), indicating that all of the metrics tested were strongly discriminative of authorship. The shared rare terms metric had by far the best discrimination ratio but the worst coverage, indicating that two title+abstracts are unlikely to share any rare terms at all even when they are written by the same individual, but this is extremely unlikely when they are written by different individuals. In contrast, the longest common character string metric had complete coverage (every article pair shared at least 3 common characters) but had the lowest discrimination ratio; even article pairs written by different individuals could share long multi-word strings such as "in the context of global climate change". The number of shared title words metric fell in between the other two direct metrics; only about half of the positive article pairs shared one or more title terms, but less than $1/20^{th}$ of the negative pairs shared any title terms at all. The three implicit metrics exhibited a nice balance of high coverage (97.4% - 100%) and high discrimination ratios (16.08 -21.27).

In order to learn whether the different metrics were capturing distinct features or were largely redundant on this task, we compared the Spearman rank correlations of the similarity scores of each metric against each other. For each metric, the different article pairs across the positive sample will naturally vary considerably in terms of their similarity. This is true too to some extent for pairs contained in the negative sample. However, since the overall similarities are so low in the negative sample, the range of variation and the shape of the distribution of scores is likely to be quite different in the positive vs. the negative samples. Thus, this analysis was studied separately for the positive sample, for the negative sample, and for the two samples combined together (which will exhibit the largest overall range of similarity scores).

As shown in Table 9, the implicit shared terms and implicit weighted scores were highly correlated (>0.95) in all cases, but were much less correlated with the word2vec-based pvtopic score (rank correlations ranging from 0.3255 in the negative sample to 0.7866 in the combined sample). This indicates that the implicit term metrics introduced in this paper are identifying different aspects of textual similarity than pvtopic, even though both types of metrics showed similar performance at discriminating authorship. All three of the implicit metrics showed limited correlations with the direct metrics (~0.5 – ~0.7 in the combined sample), and the direct metrics showed similar limited correlations amongst themselves as well.

**DISCUSSION**

In the present paper, we have utilized PubMed titles and abstracts as a corpus to compute several term and text passage similarity metrics, which have been characterized and compared in detail, and are presented as datasets that can be publicly queried or downloaded for further use. Although our interest is focused on biomedical applications, the metrics can be computed for any large textual corpus and should have general applicability in text mining across domains.

Three types of term similarity metrics were studied in this paper: a) the **direct odds ratio**, which measures how often two words are observed to co-occur in the same article, relative to the value that would be expected by chance; b) **word2vec** similarity, which represents each word as a neural embedding vector and measures similarity as the cosine similarity of any two vectors; and c) our novel unweighted and weighted implicit similarity scores, **implicit shared terms** and **implicit weighted score**, which represent each term as a vector consisting of their 300 most similar context terms (ranked according to direct similarity. We also studied text passage similarity metrics consisting of: a) the **pvtopic** formulation of text similarity computed using paragraph2vec [26], and b) our novel unweighted and weighted **implicit weighted score** measures, which represent each title+abstract as a vector consisting of their 300 most similar context terms (according to direct similarity), and compute the similarity of any two text passages by counting how many terms are shared in their vector representations.

The major methodological innovation introduced here is the way we have formulated the term and text vector representations and computed their similarities. To our knowledge, no other term or text similarity metrics combine these desirable features: 1. They are unsupervised methods, which require no training, have a minimum of tunable parameters, and are readily computed from direct co-occurrence data. 2. They represent words, phrases, or text as a low dimensional vector, in which the meaning and relative importance of dimensions is transparent to inspection. 3. They measure implicit similarity, i.e., the sharing of similar context words and phrases. The most similar system to the present one is Liu et al [36] who also represented documents as an implicit similarity vector. In their scheme, keyphrases (rather than text words and phrases) are extracted from text, and each keyphrase is represented as a weighted vector ('silhouette') consisting of co-occurring keyphrases. After pruning, the optimal number of dimensions in this vector is similar to that studied here, i.e., 200-400. Like our system, their method is transparent and interpretable, and uses implicit similarity measures. However, our representation of a document emerges naturally from its words and phrases, which theirs does not. Also, unlike our scheme, they a) compute a single ranking of latent keywords across the entire corpus of documents, b) they use a different weighting scheme that requires extensive training to assign the weights, and c) they compute the similarity of two documents as the cosine similarity of their vectors.

When choosing a similarity metric for a given task on a given corpus, there exist several considerations. Certainly, one would like a metric that exhibits high performance and is easy to compute. As well, two other factors may be less obvious. One is generalizability of the metric [28, 29]: Does the metric give "reasonably" high performance on a wide range of tasks and corpora without the need for custom tuning? Another is the extent to which one metric to be employed in modeling text (e.g., classification or clustering tasks) is redundant with other features already incorporated in the model. The direct odds ratio, implicit shared terms, and implicit weighted score all gave similar high performance on a variety of biomedical term similarity benchmarks, in some cases higher than any other metrics that have been examined to date where the pair of terms is the sole input to the metric (i.e., excluding approaches which employ supplementary outside information from knowledgebases) [e.g., 5, 9, 27, 28]. These all showed relatively good results without any need for custom tuning. Both the direct odds ratio and implicit shared terms are easy to compute. Yet they clearly capture somewhat different aspects of similarity (Tables 1 and 2) and are not truly redundant in either term similarity / relatedness tasks (Tables 3-7) or article similarity rankings (Tables 8 and 9). The implicit term metrics can be thought of as a generalization or smoothing of the direct odds ratio, providing better sampling (and thus more robustness) when direct co-occurrences are low. Also, the implicit text metrics are more sensitive than direct similarity metrics, since it can provide a measure of similarity even when there are no shared terms at all. Thus, we suggest that the novel weighted and unweighted implicit similarity metrics presented here may be uniquely valuable for biomedical text modeling.

We also compared the direct odds ratio, implicit shared terms, and implicit weighted score metrics against word2vec-based term and text passage metrics. Word2vec appeared to have inferior performance on the biomedical term similarity / relatedness benchmarks, as assessed both by our tests and those reported by others (see Results), notwithstanding the fact that word2vec performance can be optimized for specific tasks by adjusting different choices of parameters and hyperparameters and different ways of handling multi-word phrases [28, 34]. More importantly, word2vec and pvtopic (the word2vec-based implementation employed for text) clearly capture somewhat different aspects of similarity than the implicit metrics introduced here.

Apart from author name disambiguation, what other types of text modeling tasks might benefit from using our implicit term and text similarity metrics? Word2vec-based metrics have been very actively explored for a variety of extrinsic tasks such as named entity recognition, part of speech tagging, ranking PubMed articles for semantic relatedness [18, 26], and word sense disambiguation [37]. One should be cautioned that performance on one set of tasks does not necessarily correlate with performance on other tasks [38, 39], yet we feel that our implicit term similarity metrics should be explored to see if they have any complementary value in modeling the latter applications.

**IMPLEMENTATION**

In order to foster further research on biomedical term similarity and article similarity metrics, we have provided free, public access to our datasets on the Arrowsmith project website http://arrowsmith.psych.uic.edu/arrowsmith_uic/word_similarity_metrics.html. These data are being released under the terms of the Creative Commons Attribution-NonCommercial-ShareAlike CC BY-NC-SA International Public License 4.0. Specifically:

1. The basic word model, consisting of word vector representations and pre-computed word-word similarity measures, is available as datasets that can be downloaded.

2. The full term model, consisting of term vector representations for words, abbreviations, bigrams, and trigrams, and pre-computed term-term similarity measures, is available as datasets that can be downloaded.

3. All PubMed articles (having English abstracts, for which the title+abstract contains at least 25 words) have been represented as 300-dimensional vectors by both the implicit weighted score and by pvtopic. The vector representations are available for download, and as new articles appear weekly, we plan to incrementally add new vectors to the dataset.

3. A query interface at http://arrowsmith.psych.uic.edu/cgi-bin/arrowsmith_uic/word_metrics.cgi permits the user to enter any word or term (from either the basic or full model) and view the 300 most similar words or terms as pre-computed according to the direct odds ratio, implicit shared terms, implicit weighted score, and word2vec. The interface also shows normalized values by giving the percentile value corresponding to each score (e.g., a given similarity score may be greater than 98% of all similarity scores in the dataset).

4. Another query interface at http://arrowsmith.psych.uic.edu/cgi-bin/arrowsmith_uic/doc_sim_scoring2.cgi permits the user to enter any two PubMed article IDs (PMIDs) and view the weighted and unweighted implicit similarity scores. Note that the unweighted implicit shared terms scores range from 0 to 300.

5. Finally, a query interface at http://arrowsmith.psych.uic.edu/cgi-bin/arrowsmith_uic/pvtopic_sim.cgi permits the user to enter any two PubMed article IDs and view the pvtopic similarity scores. Note that the pvtopic similarity scores range from -1 to +1.

## ACKNOWLEDGEMENTS

Our studies are supported by NIH grants R01LM10817 and P01AG03934. We thank Ruixue Wang (Wuhan University) for computing some of the word2vec word similarity scores. Thanks, too, to Keven Bretonnel Cohen for suggesting a more vivid introductory paragraph.

**Table 1. Top 10 words most related to "Tennis" according to three metrics.**

| Rank | Direct Odds Ratio | Implicit Weighted Score | Word2vec |
|---|---|---|---|
| 1 | player | soccer | basketball |
| 2 | ball | athletic | volleyball |
| 3 | elbow | basketball | soccer |
| 4 | epicondylitis | athlete | badminton |
| 5 | sport | sport | handball |
| 6 | athlete | throwing | baseball |
| 7 | elite | volleyball | football |
| 8 | tournament | baseball | rugby |
| 9 | soccer | football | softball |
| 10 | golf | collegiate | amateur |

**Table 2. Top 10 words most related to "Italy" according to three metrics.**

| Rank | Direct Odds Ratio | Implicit Weighted Score | word2vec |
|---|---|---|---|
| 1 | emilia | spain | france |
| 2 | tuscany | europe | spain |
| 3 | veneto | italian | greece |
| 4 | sardinia | france | portugal |
| 5 | campania | greece | romania |
| 6 | lombardy | portugal | belgium |
| 7 | romagna | european | germany |
| 8 | sicily | poland | turkey |
| 9 | turin | inhabitant | switzerland |
| 10 | milan | mediterranean | sweden |

**Table 3. Spearman rank correlations among human relatedness scores and similarity metrics for the 30 biomedical term relatedness benchmark.**

|  | Physician | Coder | Direct Odds Ratio | Imp Shared | Imp Weighted | Best word2vec | Best pvtopic |
|---|---|---|---|---|---|---|---|
| Physician | 1.0000 | 0.8886 | 0.8425 | 0.8611 | 0.8597 | 0.7926 | 0.6759 |
| Coder |  | 1.0000 | 0.8247 | 0.8103 | 0.8127 | 0.8008 | 0.6924 |
| Direct Odds Ratio |  |  | 1.0000 | 0.9133 | 0.9047 | 0.7516 | 0.8282 |
| Implicit Shared Terms |  |  |  | 1.0000 | 0.9781 | 0.7644 | 0.8650 |
| Implicit Weighted Score |  |  |  |  | 1.0000 | 0.7749 | 0.8630 |
| Best word2vec |  |  |  |  |  | 1.0000 | 0.6605 |
| Best pvtopic |  |  |  |  |  |  | 1.0000 |

Implicit shared terms and implicit weighted score are computed from the full term model (that includes words as well as selected bigrams, trigrams, and abbreviations). Shown are the versions of word2vec and pvtopic that gave the best performance among those tested, namely, summing vectors to represent phrases, and computing similarity on exact words used in the benchmarks, rather than on the terms that mapped to our dataset (Supplemental File 1).

**Table 4. Spearman rank correlations among human relatedness scores and similarity metrics for the 101 biomedical term relatedness benchmark.**

|  | Mean | Direct Odds | Implicit Shared | Implicit Weighted | word2vec |
|---|---|---|---|---|---|
| Mean Relatedness Judgment | 1.0000 | 0.7582 | 0.7426 | 0.7354 | 0.4968 |
| Direct Odds Ratio |  | 1.0000 | 0.8673 | 0.8703 | 0.5539 |
| Implicit Shared Terms |  |  | 1.0000 | 0.9879 | 0.6823 |
| Implicit Weighted Score |  |  |  | 1.0000 | 0.6837 |
| word2vec |  |  |  |  | 1.0000 |

**Table 5. Spearman rank correlations among human similarity scores and similarity metrics for the UMNSRS-Similarity benchmark (evaluating the 501 term pairs that mapped to our dataset).**

|  | Mean | Direct Odds Ratio | Imp Shared | Imp Weighted | word2vec |
|---|---|---|---|---|---|
| Mean Similarity Judgement | 1.0000 | 0.7013 | 0.6649 | 0.6931 | 0.5797 |
| Direct Odds Ratio |  | 1.0000 | 0.8322 | 0.8566 | 0.7387 |
| Implicit Shared Terms |  |  | 1.0000 | 0.9826 | 0.8194 |
| Implicit Weighted Score |  |  |  | 1.0000 | 0.8236 |
| word2vec |  |  |  |  | 1.0000 |

**Table 6. Spearman rank correlations among medical student relatedness scores and similarity metrics for the UMNSRS-Relatedness benchmark (evaluating the 511 term pairs that mapped to our dataset).**

|  | Mean | Direct Odds Ratio | Imp Shared | Imp Weighted | word2vec |
|---|---|---|---|---|---|
| Mean Similarity Judgement | 1.0000 | 0.6338 | 0.5645 | 0.5973 | 0.4832 |
| Direct Odds Ratio |  | 1.0000 | 0.7998 | 0.8272 | 0.7197 |
| Implicit Shared Terms |  |  | 1.0000 | 0.9819 | 0.8229 |
| Implicit Weighted Score |  |  |  | 1.0000 | 0.8261 |
| word2vec |  |  |  |  | 1.0000 |

**Table 7. Spearman rank correlations among medical student relatedness scores and similarity metrics for the modified UMNSRS-Relatedness benchmark (evaluating the 420 term pairs that mapped to our dataset).**

|  | Mean | Direct Odds Ratio | Imp Shared | Imp Weighted | word2vec |
|---|---|---|---|---|---|
| Mean Similarity Judgement | 1.0000 | 0.6658 | 0.5892 | 0.6226 | 0.5062 |
| Direct Odds Ratio |  | 1.0000 | 0.7867 | 0.8129 | 0.6992 |
| Implicit Shared Terms |  |  | 1.0000 | 0.9828 | 0.8137 |
| Implicit Weighted Score |  |  |  | 1.0000 | 0.8125 |
| word2vec |  |  |  |  | 1.0000 |

**Table 8. Article similarity metrics computed for pairs of sole-author articles written by the same individual (positive sample) vs. written by different individuals (negative sample).**

**Positive Sample:**

|  | shared title words | longest common string | rare terms | implicit shared terms | implicit weighted score | pvtopic |
|---|---|---|---|---|---|---|
| **Mean:** | 1.26 | 20.64 | 3.77 | 150.91 | 2328.80 | 0.18 |
| **Std. Dev.:** | 1.83 | 14.50 | 10.47 | 102.41 | 2312.15 | 0.11 |
| **# nonzero values** | 508 | 1000 | 292 | 974 | 974 | 1000 |

**Negative Sample:**

|  | shared title words | longest common string | rare terms | implicit shared terms | implicit weighted score | pvtopic |
|---|---|---|---|---|---|---|
| **Mean:** | 0.05 | 9.06 | 0.00 | 9.05 | 109.48 | 0.01 |
| **Std. Dev.:** | 0.24 | 2.55 | 0.03 | 21.61 | 315.56 | 0.07 |
| **# nonzero values** | 48 | 1000 | 1 | 426 | 426 | 1000 |
| **Discrim Ratio** | 24.65 | 2.28 | 3774 | 16.67 | 21.27 | 16.08 |

All metrics were significantly different (positive vs. negative sample) at p<0.0001, using 2-tailed t-test, unpaired.

**Table 9. Spearman rank correlations among similarity metrics in pairs of sole-author articles written by the same individual (positive sample) vs. written by different individuals (negative sample) and for the two sets combined.**

**Positive Sample:**

|  | implicit shared terms | implicit weighted score | shared title words | longest common string | rare terms | pvtopic |
|---|---|---|---|---|---|---|
| **implicit shared terms** | 1.0000 | 0.9548 | 0.5148 | 0.6146 | 0.4720 | 0.6424 |
| **implicit weighted score** |  | 1.0000 | 0.4778 | 0.5916 | 0.4325 | 0.6219 |
| **shared title words** |  |  | 1.0000 | 0.5909 | 0.5594 | 0.5393 |
| **longest common string** |  |  |  | 1.0000 | 0.7200 | 0.6326 |
| **rare terms** |  |  |  |  | 1.0000 | 0.5858 |
| **pvtopic** |  |  |  |  |  | 1.0000 |

**Negative Sample:**

|  | implicit shared terms | implicit weighted score | shared title words | longest common string | rare terms | pvtopic |
|---|---|---|---|---|---|---|
| **implicit shared terms** | 1.0000 | 0.9979 | 0.1190 | 0.2129 | 0.0307 | 0.3270 |
| **implicit weighted score** |  | 1.0000 | 0.1189 | 0.2232 | 0.0254 | 0.3255 |
| **shared title words** |  |  | 1.0000 | 0.1506 | -0.0071 | 0.0912 |
| **longest common string** |  |  |  | 1.0000 | -0.0446 | 0.1566 |
| **rare terms** |  |  |  |  | 1.0000 | 0.0464 |
| **pvtopic** |  |  |  |  |  | 1.0000 |

**Combined:**

|  | implicit shared terms | implicit weighted score | shared title words | longest common string | rare terms | pvtopic |
|---|---|---|---|---|---|---|
| **implicit shared terms** | 1.0000 | 0.9923 | 0.6130 | 0.7081 | 0.5093 | 0.7866 |
| **implicit weighted score** |  | 1.0000 | 0.5994 | 0.7054 | 0.4943 | 0.7795 |
| **shared title words** |  |  | 1.0000 | 0.5959 | 0.5806 | 0.5906 |
| **longest common string** |  |  |  | 1.0000 | 0.5840 | 0.6644 |
| **rare terms** |  |  |  |  | 1.0000 | 0.5341 |
| **pvtopic** |  |  |  |  |  | 1.0000 |

**Supplemental Material**

**Supplemental File 1.** The 30 term relatedness benchmark [ref], showing the 30 terms, their mapping to our dataset, and the relatedness ratings of physicians and coders (obtained from http://rxinformatics.umn.edu/SemanticRelatednessResources.htm). Also shown for each term pair are the similarity values computed according to 5 different metrics: The direct odds ratio, the implicit shared terms, implicit weighted score, word2vec and pvtopic. The latter two metrics are computed in a variety of ways (either using the exact terms or the mapped terms, and either using single word vectors or summed vectors). The Spearman rank correlations are shown for all metrics relative to physician and coder relatedness ratings and among themselves.

**Supplemental File 2.** The 101 term relatedness benchmark [ref], showing the terms, mapping to our dataset, and the relatedness ratings of coders (obtained from http://rxinformatics.umn.edu/SemanticRelatednessResources.htm). One term pair failed to map and was excluded, giving a total of 100 word pairs to evaluate. Also shown for each term pair are the similarity values computed according to 4 different metrics: The direct odds ratio, the implicit shared terms, implicit weighted score, and word2vec.

**Supplemental File 3.** The UMNSRS-Similarity benchmark [ref], showing the 501 term pairs that mapped to our dataset, and the similarity ratings of coders (obtained from http://rxinformatics.umn.edu/SemanticRelatednessResources.htm). Also shown for each term pair are the similarity values computed according to 4 different metrics: The direct odds ratio, the implicit shared terms, implicit weighted score, and word2vec.

**Supplemental File 4.** The UMNSRS-Relatedness benchmark [ref], showing the 511 term pairs that mapped to our dataset, and the mean +/- SD similarity ratings of medical students (obtained from http://rxinformatics.umn.edu/SemanticRelatednessResources.htm). Also shown for each term pair are the similarity values computed according to 4 different metrics: The direct odds ratio, the implicit shared terms, implicit weighted score, and word2vec.

**Supplemental File 5.** The modified UMNSRS-Relatedness benchmark [ref], showing the 420 term pairs that mapped to our dataset, and the mean similarity ratings of medical students (obtained from http://rxinformatics.umn.edu/SemanticRelatednessResources.htm). Also shown for each term pair are the similarity values computed according to 4 different metrics: The direct odds ratio, the implicit shared terms, implicit weighted score, and word2vec.

**Supplemental File 6.** Pairs of PubMed sole-authored articles, including 1,000 pairs randomly chosen that are predicted to be written by the same individual (positive set on Sheet 1), and 1,000 pairs that are predicted to be written by different individuals (negative set on Sheet 2). Shown the pairs of PMIDs, and their direct and implicit similarity scores as computed according to six different metrics (see text). Sheet 3 shows the mean and SD values of the similarity scores in each set, and sheet 4 shows the Spearman rank correlation values among the different metrics.